\RequirePackage{fix-cm}
\documentclass[twocolumn,envcountsect,envcountsame]{svjour3}  
\usepackage{graphicx}
\usepackage{footmisc}
\usepackage[numbers, sort, comma, square]{natbib}
\usepackage{amsmath}
\usepackage{bm}
\usepackage{amsfonts}  
\usepackage{nicefrac}  
\usepackage{microtype}  
\usepackage{lipsum}
\usepackage{tabularx}
\usepackage{bm}
\usepackage{multicol} 
\usepackage{multirow}
\usepackage{threeparttable}
\usepackage{float}
\restylefloat{table}
\usepackage{lipsum}
\usepackage[linesnumbered,ruled,vlined]{algorithm2e}
\usepackage{hyperref}
\usepackage[switch]{lineno}
\usepackage{mathtools, cuted}
\usepackage{amsthm}
\usepackage{amsmath,amsfonts,bm}

\makeatletter
\def\BState{\State\hskip-\ALG@thistlm}
\makeatother

\def\eqref#1{equation~\ref{#1}}

\def\1{\mathbb{I}}

\def\rw{{\textnormal{w}}}

\def\ry{{\textnormal{y}}}

\def\rx{{\textnormal{x}}}

\def\rvb{{\mathbf{b}}}

\def\rvu{{\mathbf{i}}}

\def\rvu{{\mathbf{u}}}
\def\rvv{{\mathbf{v}}}

\def\rvz{{\mathbf{z}}}

\def\rvx{{\mathbf{x}}}

\def\rmA{{\mathbf{A}}}

\def\vb{{\bm{b}}}

\def\vu{{\bm{u}}}

\def\vz{{\bm{z}}}

\def\vx{{\bm{x}}}

\def\mA{{\bm{A}}}
\def\mB{{\bm{B}}}
\def\mC{{\bm{C}}}

\def\mI{{\bm{I}}}

\DeclareMathAlphabet{\mathsfit}{\encodingdefault}{\sfdefault}{m}{sl}
\SetMathAlphabet{\mathsfit}{bold}{\encodingdefault}{\sfdefault}{bx}{n}

\def\sH{{\mathcal{H}}}

\def\sN{{\mathcal{N}}}

\def\sX{{\mathcal{X}}}
\def\sY{{\mathcal{Y}}}
\def\sZ{{\mathcal{Z}}}

\newcommand{\E}{\mathbb{E}}

\newcommand{\R}{\mathbb{R}}

\newcommand\independent{\protect\mathpalette{\protect\independenT}{\perp}}
\def\independenT#1#2{\mathrel{\rlap{$#1#2$}\mkern2mu{#1#2}}}

\spnewtheorem{thm}{Theorem}[section]{\bfseries}{\itshape}
\spnewtheorem{lem}[thm]{Lemma}{\bfseries}{\itshape}
\spnewtheorem{dfn}[thm]{Definition}{\bfseries}{\itshape}
\spnewtheorem{cor}[thm]{Corollary}{\bfseries}{\itshape}

\smartqed
\begin{document}

\title{Effective Sample Size, Dimensionality, and Generalization in Covariate Shift Adaptation
}


\author{Felipe Maia Polo    \and
  Renato Vicente 
}


\institute{Felipe Maia Polo \at
    Department of Statistics, Institute of Mathematics and Statistics of the University of São Paulo \\ Advanced Institute for Artificial Intelligence (AI2) \\
    São Paulo, São Paulo, Brazil \\
    ORCID: 0000-0002-4950-2795\\
    \email{felipemaiapolo@gmail.com}   
   \and
   Renato Vicente \at
    Department of Applied Mathematics, Institute of Mathematics and Statistics of the University of São Paulo \\ Experian DataLab LatAm  \\ Advanced Institute for Artificial Intelligence (AI2)\\
    São Paulo, São Paulo, Brazil \\
    ORCID: 0000-0003-0671-9895\\
    \email{rvicente@usp.br} 
}

\date{}

\maketitle

\begin{abstract}
In supervised learning, training and test\linebreak datasets are often sampled from distinct distributions. Domain adaptation techniques are thus required. Covariate shift adaptation yields good generalization performance when domains differ only by the marginal distribution of features. Covariate shift adaptation is usually implemented using importance weighting, which may fail, according to common wisdom, due to small effective sample sizes (ESS). Previous research argues this scenario is more common in high-dimensional settings. However, how effective sample size, dimensionality, and model performance/generalization are formally related in supervised learning, considering the context of covariate shift adaptation, is still somewhat obscure in the literature. Thus, a main challenge is presenting a unified theory connecting those points. Hence, in this paper, we focus on building a unified view connecting the ESS, data dimensionality, and generalization in the context of covariate shift adaptation.  Moreover, we also demonstrate how dimensionality reduction or feature selection can increase the ESS, and argue that our results support dimensionality reduction before covariate shift adaptation as a good practice.

\end{abstract}

\section{Introduction}\label{sec:intro}

A fundamental assumption in supervised statistical learning is that the data used to train our models and the data we want to make predictions for are sampled from the same distribution. Usually, real-world machine leaning applications, explicitly or implicitly, rely on this assumption\footnote{See, for example, \citet{ga_wu2013prediction,ga_banan2020deep,ga_cheng2020forecasting,ga_ghalandari2019aeromechanical,ga_fan2020spatiotemporal,ga_taormina2015ann}}. However, that assumption is violated when there is covariate shift \cite{shimodaira2000improving,sugiyama2012machine}. In this scenario, we have a training/source joint distribution $Q_{\rvx,\ry}$ which differs from the test/target distribution $P_{\rvx,\ry}$. Features are sampled from different marginals $Q_{\rvx}\neq P_{\rvx}$ while labels are sampled according to the same conditional distribution $Q_{\ry|\rvx}=P_{\ry|\rvx}$. In the training phase, labeled pairs $\{(\rvx_i, \ry_i)\}_{i = 1}^n$ are identically and independently sampled from $Q_{\rvx,\ry}$, while unlabeled vectors $\{\rvx'_i\}_{i = 1}^m $ are identically and independently sampled from $P_{\rvx}$. If the marginal distributions of features have density functions $p_{\rvx}$ and $q_{\rvx}$, such that $\text{support}(p_{\rvx}) \subseteq \text{support}(q_{\rvx})$, the most common approach to adapt a model for the target distribution  is to employ an empirical error weighted by $w(\vx)=p_{\rvx}(\vx)/q_{\rvx}(\vx)$ \cite{shimodaira2000improving,huang2007correcting,sugiyama2007covariate,kanamori2009least,sugiyama2012machine}. 

The weighting scheme may fail when small effective sample sizes (ESS) are small. According to common wisdom, a small ESS hurts model’s performance in the target distribution. As previous research argues, e.g., \citet{wang2017extreme}, that kind of scenario is common when working with high-dimensional data. However, to the best of our knowledge, there is no unified and rigorous view on how the three key concepts (i) effective sample size (ESS), (ii) data dimensionality, and (iii) generalization of supervised models under covariate shift are connected to each other. In this paper, we present a unified theory connecting the three concepts. Moreover, we also explore how dimensionality reduction or feature selection can increase the effective sample size. 

This paper is organized as follows. In Section \ref{sec:rel_work}, we discuss previous results and explain our contribution to the debate. In Section \ref{sec:ess}, we briefly review importance weighting and introduce a new connection between effective sample size and generalization in the context of covariate shift adaptation. In Section \ref{sec:dim}, we introduce dimensionality to the problem showing how it connects to the other two concepts and then illustrate these connections with a toy experiment. Finally, in Section \ref{sec:dim_red}, we show how dimensionality reduction and feature selection can lead to a larger effective sample size. We conclude our discussion with real-data experiments that supports feature selection before covariate shift adaptation as a good practice.

\section{Related Work}\label{sec:rel_work}

There is a rich literature on the problem of covariate shift adaptation\footnote{See \citet{sugiyama2012machine} for a general view.} or related subjects. The main interest has been to develop methods to estimate the density ratio $w$ \cite{huang2007correcting,sugiyama2008direct,kanamori2009least,izbicki2014high,liu2017trimmed}. Some of the proposed methods aim to reliably estimate $w$ in high-dimensional and unstable settings \cite{izbicki2014high,liu2017trimmed}, when the more traditional approaches may fail. However, according to the common wisdom of the area, even if we could perfectly estimate $w$, we would still have to deal with poor performance due to small effective sample sizes (ESS), especially in high-dimensional settings. Understanding the role of small ESS and possible ways to attenuate it may, therefore, be productive. The covariate shift adaptation literature has already tried to articulate the relationships between ESS and generalization in high-dimensional settings, also proposing dimensionality reduction as a cure. In spite of that, we believe these previous attempts fail in connecting these concepts in a unified manner and as explicitly as we propose to do in this paper.  

In recent years, \citet{reddi2015doubly} proposed a regularization method that controls the ESS and offers sharper generalization bounds while correcting for covariate shift. However, the authors do not explore how the number of features plays an essential role. Another work that explores the concept of ESS in the context of covariate shift adaptation is \citet{gretton2009covariate}. In that work, the authors present the relationship between ESS and generalization bounds in a transductive learning scenario. Besides transductive learning not being as common as inductive learning in practice, the authors also do not explore how dimensionality plays an essential role in the problem.

The idea of features dimensionality being related to ESS is  explored in \citet{wang2017extreme}, without formalizing the connection to generalization. The authors also motivate how dimensionality reduction can make ESS bigger, however, the central hypothesis adopted in this case is that dimensionality reduction does not depend on the data, which, in most cases, is not valid. In a more recent paper, \citet{stojanov2019low} proposes a dimensionality reduction method to make covariate shift adaptation feasible, especially when estimating weights.  The authors show how the number of features is indirectly related to transductive generalization bounds and effective sample size when the correction is made by Kernel Mean Matching \cite{huang2007correcting}. In addition to the results being restricted to a particular case, the authors implicitly assume that the mapping that defines dimensionality reduction is given beforehand and does not depend on the training data, what is not realistic.

In this paper, we complement previous works by formally articulating the relationship among ESS, generalization of predictive models in the inductive scenario, and dimensionality as explicitly as possible. We present a unified theory connecting the three concepts, which was not observed by us in the literature. We also show that dimensionality reduction, even considering that the mapping may depend on the data, mitigates low ESS by making the source and target domains less divergent.


\section{Effective Sample Size (ESS) and Generalization in Covariate Shift Adaptation}\label{sec:ess}

\subsection{Importance Weighting}

To keep our discussion as self-contained as possible, we first use this subsection to quickly summarize key points behind importance weighting.

Given a hypothesis class $\mathcal{H}$ and a loss function $L$, our goal is finding a hypothesis $h^* \in \sH$ that minimizes the risk $R$ assessed in the target distribution $P_{\rvx,\ry}$ using data from source distribution $Q_{\rvx,\ry}$. From now on we assume: (i) $Q_{\ry|\rvx}=P_{\ry|\rvx}$ and $Q_{\rvx}\neq P_{\rvx}$; (ii) distributions $P_{\rvx}$ and $Q_{\rvx}$ have probability density functions (p.d.f.s) $p_{\rvx}$ and $q_{\rvx}$ such that $\text{support}(p_{\rvx}) \subseteq \text{support}(q_{\rvx})$.  Then, the risk can be written in terms of the source distribution:
\begin{align}
 R(h)&= \E_{\rvx \sim P_{\rvx}} \E_{\ry|\rvx} \left[L(h(\rvx),\ry) \right] \\[.5em]
 &= \int \frac{p_{\rvx}(\vx)}{q_{\rvx}(\vx)} q_{\rvx}(\vx) \E_{\ry|\rvx} \left[L(h(\vx),\ry) \right] \text{d}\vx \\[.5em]
 &= \E_{\rvx \sim Q_{\rvx}} \E_{\ry|\rvx} \left[w(\rvx) \cdot L(h(\rvx),\ry) \right]
\end{align}

We would like to find a hypothesis $h^{\text{ERM}}_{\hat{w}} \in \mathcal{H}$ that minimizes a weighted version of the empirical risk while also obtaining a low value for $R$. Assume we have an estimate $\hat{w}$ for the “true" weighting function $w=p_{\rvx}/q_{\rvx}$ and that we have pairs $\{(\rvx_i,\ry_i)\}_{i=1}^n$ that are identically and independently (i.i.d.) sampled from $Q_{\rvx,\ry}$. The weighted empirical risk is thus given by

\begin{align}
 \widehat{R}_{\hat{w}}(h)&=\frac{1}{n}\sum_{i=1}^{n} \hat{w}(\rvx_i) \cdot L(h(\rvx_i),\ry_i)
\end{align}

In practice, we might also want to add a regularization term $\Omega(h)$ to penalize for the complexity of the hypothesis $h$.

\subsection{Relationship of Effective Sample Size (ESS) and Generalization in Covariate Shift Adaptation}

To introduce the concept of effective sample size in the context of covariate shift adaptation, we first describe how this heuristic is employed within the importance sampling literature \cite{robert2010introducing,mcbook,martino2017effective}, where it originally comes from. We assume the “true” importance function (density ratio) is known up to a constant. This assumption enables us to achieve some theoretical results and is also adopted in previous works \cite{wang2017extreme,cortes2010learning,cortes2019relative}. The strategy we use to show the relevance of the effective sample size in covariate shift adaptation is to find an asymptotic approximation for that quantity, and then connect it to a known generalization bound.

The ESS formulation we use is slightly different from the most usual one \cite{robert2010introducing,mcbook,martino2017effective} in the sense we are concerned with percentage of effective samples and not with the number of effective samples\footnote{In the literature, it is common to present the ESS as $n\cdot\widehat{\textup{ESS}}_n$, while we are concerned only with $\widehat{\textup{ESS}}_n$ (Equation \ref{defn:ess}).}. Given the two definitions are not very different, the intuitions and some results regarding ESS are easily adaptable. We present our definition in the following.

Consider two probability distributions $P_{\rvz}$ and $Q_{\rvz}$ over $\sZ \subseteq \R^d$ with probability density functions $p_{\rvz}$ and $q_{\rvz}$ such that $\text{support}(p_{\rvz}) \subseteq \text{support}(q_{\rvz})$. From now on, we call $P_{\rvz}$ the \textit{target} distribution and $Q_{\rvz}$ the \textit{source} distribution. We thus sample from $Q_{\rvz}$ in order to estimate the integral $\int_\sZ g(\vz) p_{\rvz}(\vz) d\vz=\int_\sZ \frac{p_{\rvz}(\vz)}{q_{\rvz}(\vz)} g(\vz) q_{\rvz}(\vz) d\vz$, with $g: \sZ \rightarrow \R$ integrable. A key quantity in this problem is the importance function, which is given by $w \propto p_{\rvz}/q_{\rvz}$. 

Suppose we have an independent and identically distributed (i.i.d.) sample $\{\rvz_i\}_{i=1}^n$ from the \text{source} distribution $Q_{\rvz}$ and we want to use the (self-normalized\footnote{We show the case of the self-normalized estimator because it returns the most usual definition for the ESS, which is also used in the context of covariate shift \cite{reddi2015doubly}. In spite of that, we show that this definition for the ESS is still useful for the non normalized case while performing covariate shift adaptation.}) importance sampling estimator $n^{-1}\sum_{i=1}^n \bar{\rw}_i g(\rvz_i)$ in order to estimate the integral of interest. The weights are given by $\bar{\rw}_i=\rw_i/\sum_j \rw_j$, where $\rw_i =w(\rvz_i) \propto p_{\rvz}(\rvz_i)/q_{\rvz}(\rvz_i)$, $i \in [n]:=\{1,...,n\}$. Then, the effective sample size is defined as
\begin{align}\label{defn:ess}
 \widehat{\textup{ESS}}_n(P_\rvz, Q_\rvz)&:=\frac{1}{n\sum_{i=1}^n \bar{\rw}_i^2}\\[.5em]
 &=\frac{(\sum_{i=1}^n \rw_i)^2}{n\sum_{i=1}^n \rw_i^2}
\end{align}

Intuitively, the effective sample size is the percentage of effective samples. For example, if the effective sample size equals $1/2$, then the importance sampling estimator \textit{effectiveness} is the same of a monte carlo estimator with $n/2$ samples. That formulation can be used to approximate, via Delta Method, the ratio of monte carlo estimators' variance and the self-normalized importance sampling estimator' variance, using the derivation made by \citet{elvira2018rethinking}. While that work motivates the use of the ESS, other approaches can be derived from \citet{mcbook} and \citet{martino2017effective}. The latter presents the relationship between effective sample size and the euclidean distance between the vector $(\bar{\rw}_1,...,\bar{\rw}_n)$ and the “ideal" balanced vector $(1/n,...,1/n)$. Furthermore, effective sample size informs about the importance sampling estimator's convergence rate \cite{agapiou2017importance}. Said that, the results presented in this section for the covariate shift adaptation case resembles the results presented by \citet{agapiou2017importance} in a different context.

To move forward, we introduce the concept of Rényi Divergence, which plays a central role in our analysis:

\begin{dfn}[Rényi Divergence \cite{van2014renyi}]
Consider two probability distributions $P_{\rvx}$ and $Q_{\rvx}$ over $\sX \subseteq \R^d$, with probability density functions $p_{\rvx}$ and $q_{\rvx}$ such that $\text{support}(p_{\rvx}) \subseteq \text{support}(q_{\rvx})$. The Rényi Divergence of order $\alpha>1$ of $P_{\rvx}$ from $Q_{\rvx}$ is given by:
\begin{align}
  D_\alpha(P_\rvx || Q_\rvx):=\frac{1}{\alpha-1}\log \underset{\rvx\sim Q_{\rvx}}{\E}\Bigg[\left(\frac{p_{\rvx}(\rvx)}{q_{\rvx}(\rvx)}\right)^\alpha\Bigg]
\end{align}

Consequently, the Rényi Divergence of order 2 of \linebreak $P_\rvx$ from $Q_\rvx$ is given by $D_2(P_\rvx || Q_\rvx)=\log \E_{\rvx\sim P_{\rvx}}[\frac{p_{\rvx}(\rvx)}{q_{\rvx}(\rvx)}]$.
\end{dfn}

Despite all previous work, the question of how we should transpose the effective sample size concept to the covariate shift adaptation framework remains. In the following, we make explicit the close relationship between the ESS and generalization bounds under covariate shift adaptation. As we start talking about covariate shift adaptation, we substitute $\rvz$ by a vector of features $\rvx$, the set $\sZ$ by $\sX$ or $\sX \times \sY$ and the function $g$ by the loss function $L$. Before we move on, we must establish that the effective sample size $\widehat{\textup{ESS}}_n(P_\rvx, Q_\rvx)$ converges almost surely to the quantity $\textup{ESS}^*(P_\rvx, Q_\rvx)$, which plays a central role in our analysis. $\textup{ESS}^*(P_\rvx, Q_\rvx)$ can be considered a population version for the effective sample size. From now on, we may call it by population effective sample size or only effective sample size, when it is not ambiguous.

\begin{thm}\label{thm:ess}
Consider two probability distributions $P_{\rvx}$ and $Q_{\rvx}$ over $\sX \subseteq \R^d$, with probability density functions $p_{\rvx}$ and $q_{\rvx}$ such that $\text{support}(p_{\rvx}) \subseteq \text{support}(q_{\rvx})$. Suppose we have a random sample $\{\rvx_i\}_{i=1}^n$, identically and independently sampled from the distribution $Q_{\rvx}$, and we define $\rw_i=w(\rvx_i) \propto p_{\rvx}(\rvx_i)/q_{\rvx}(\rvx_i)$. Assume that $0<\E_{\rvx \sim Q_\rvx}\left[w(\rvx)^2 \right]< \infty$. Then
\begin{align}
\widehat{\textup{ESS}}_n(P_\rvx, Q_\rvx) \xrightarrow[n \rightarrow \infty]{a.s.}\textup{ESS}^*(P_\rvx, Q_\rvx) 
\end{align}
Where
\begin{align}
\textup{ESS}^*(P_\rvx, Q_\rvx) &:= \textup{exp}\left[-D_2(P_\rvx || Q_\rvx)\right] 
\end{align}
The quantity $D_2(P_\rvx || Q_\rvx)$ is the Rényi Divergence of order 2 of $P_\rvx$ from $Q_\rvx$ \cite{van2014renyi}.
\end{thm}

The proof can be found in the appendix. It is essential to state that similar results hold for other effective sample size definitions as, for example, the one used by \citet{wang2017extreme} divided by $n$, to give the percentage of effective samples considering the non normalized weights for covariate shift adaptation.

It is fascinating how Rényi Divergence naturally emerges when working with the effective sample size. It is a crucial point to understand that, when calculating the effective sample size, we are approximating a quantity inversely proportional to the exponential of Rényi Divergence of order 2 of $P_\rvx$ from $Q_\rvx$.

Now we focus on the understanding of how effective sample size relates to generalization of adapted supervised models. For Theorem \ref{thm:gen1}, consider some conditions. Let $\sX$ denote the input space, $\sY$ the label set, and let $L: \sY^2 \rightarrow [0, 1]$ be a bounded loss function. Denote the \textit{target} distribution of features by $P_{\rvx}$ and the \textit{source} distribution of features by $Q_{\rvx}$, such that $P_{\rvx}$ is dominated by $Q_{\rvx}$. Consider $\sH$ to be the hypothesis class used by the learning algorithm and $f: \sX \rightarrow \sY$ to be the labeling function we want to learn about. We denote by $\text{Pdim}(U)$ the pseudo-dimension\footnote{A pseudo-dimension is an extension of VC Dimension for real-valued classes of functions} of a real-valued function class $U$ \cite{vidyasagar2002theory}. $\text{Pdim}$ is used here to quantify the complexity of a hypothesis class through the loss function. Finally, $R$ is the risk assessed in the target distribution $P_{\rvx}$ and $\hat{R}_w$ is the weighted empirical error calculated using the true weighting function (density ratio) and samples $\{\rvx_i\}_{i=1}^n$, identically and independently sampled from the source distribution $Q_{\rvx}$.

\begin{thm}[Adapted from \cite{cortes2010learning}]\label{thm:gen1} 
Define the function $L_h(\vx):= L[h(\vx),f(\vx)]$ and let $\sH$ be a hypothesis set such that $\text{Pdim}(\left\{L_h: h \in \sH \right\}) = p < \infty$. Assume that
$\textup{ESS}^*(P_\rvx, Q_\rvx)=\textup{exp}\left[-D_2(P_\rvx || Q_\rvx)\right]$, $D_2(P_\rvx || Q_\rvx)<\infty$, and the target/source density ratio $w> 0$. Then, for any $\delta \in (0,1)$, with probability at least $1-\delta$, we have that:
\begin{align}
  \underset{h \in \sH}{\text{sup}}&[R(h) - \hat{R}_w(h)]  \leq \\[.5em] 
 &\leq \frac{2^{\frac{5}{4}}}{\sqrt{\textup{ESS}^*(P_\rvx, Q_\rvx)}} \cdot \left[\frac{p\cdot \textup{log}\frac{2\cdot e \cdot n }{p}+ \textup{log}\frac{4}{\delta}}{n}\right]^{\frac{3}{8}}
\end{align}
\end{thm}

See \citet{cortes2010learning} for the proof, and replace $D_2$ by $\text{ESS}^*$ to get this version of the theorem. 

It is clear from Theorem \ref{thm:gen1} that $\textup{ESS}^*(P_\rvx, Q_\rvx)$ plays a fundamental role when learning $f$ from data. A larger $\textup{ESS}^*(P_\rvx, Q_\rvx)$ leads to a tighter generalization bound. Consequently, if $\widehat{\textup{ESS}}_n(P_\rvx, Q_\rvx)$ is a good approximation for $\textup{ESS}^*(P_\rvx, Q_\rvx)$, the rationale behind using effective sample size as a heuristic for diagnosis of covariate shift adaptation becomes clearer. To conclude, we should mention that \citet{cortes2019relative} shows a similar result to Theorem \ref{thm:gen1} with less assumptions, namely, assuming the existence of a labeling function $f$ and that $ w > 0$. However, we chose the form provided by \citet{cortes2010learning}, as it gives us a more straightforward expression without losing the property that is key to our approach, to say, that a larger $\textup{ESS}^*(P_\rvx, Q_\rvx)$ leads to a sharper generalization bound.

\section{The Role of Dimensionality}\label{sec:dim}

In Section \ref{sec:ess}, we showed the effective sample size's role in the context of covariate shift adaptation exploring its asymptotic relationship with generalization bounds. However, we still need to understand the role that dimensionality plays during covariate shift adaptation. In Theorem \ref{thm:div1}, we demonstrate that the Rényi Divergence of source and target distributions does not decrease with the number of features, and, consequently, the population effective sample size does not increase with the number of features, which explains potential adaptation problems for high-dimensional data. 

\begin{thm}\label{thm:div1} 
Given two joint probability distributions $P_{\rvx_1,\rvx_2}$ (target) and $Q_{\rvx_1,\rvx_2}$ (source) over $\sX \subseteq \R^d$, \linebreak $D_2(P_{\rvx_1,\rvx_2} || Q_{\rvx_1,\rvx_2})<\infty$, with joint probability density functions $p_{\rvx_1,\rvx_2}$ and $q_{\rvx_1,\rvx_2}$, such that $\text{support}(p_{\rvx_1,\rvx_2}) \subseteq \text{support}(q_{\rvx_1,\rvx_2})$, we have that
\begin{align}
D_2(P_{\rvx_1,\rvx_2} || Q_{\rvx_1,\rvx_2}) \geq D_2(P_{\rvx_1} || Q_{\rvx_1})
\end{align}
And, consequently,
\begin{align}
 \textup{ESS}^*(P_{\rvx_1}, Q_{\rvx_1}) \geq \textup{ESS}^*(P_{\rvx_1,\rvx_2}, Q_{\rvx_1,\rvx_2})
\end{align}
\end{thm}

The proof can be found in the appendix. 

Combining the results of Theorem \ref{thm:gen1} and Theorem \ref{thm:div1}, we conclude that performing covariate shift adaptation with many features may not be feasible, as we would potentially have loose generalization bounds.

Note that Theorem \ref{thm:div1} does not necessarily say that by reducing dimensions or selecting the most relevant features we will have a bigger effective sample size. Reducing dimensions or selecting features is a random process that depends on data, and we have ignored this fact so far. In Section \ref{sec:dim_red}, we consider the randomness of the dimensionality reduction or feature selection step to prove that we can increase the effective sample size by following these procedures before conducting covariate shift adaptation.

\subsection{A Toy Experiment}\label{sec:toy}

In this section, we present a toy experiment in order to illustrate the relationship between effective sample size, Rényi divergence, dimensionality, and performance of supervised methods. 

Assume there are two joint distributions of features and labels $P_\lambda$ and $Q$ with densities $p_\lambda$ and $q$, being the case that $Q$ describes the source/training population and that $P_\lambda$ describes the target/test population. Moreover, we assume we are facing the classical covariate shift problem, that is, $p_\lambda(y|\vx)= q(y|\vx)= p(y|\vx)$ but $p_\lambda(\vx)\neq q(\vx)$, plus the fact that we cannot sample the labels from the test population. Finally, consider $q(\vx)=\sN(\vx | \mathbf{0},\mI_d)$ and $p_\lambda(\vx)=\sN(\vx | \lambda \cdot \mathbf{1},\mI_d)$, for $\lambda \neq 0$, with $d$ indicating the number of dimensions. Suppose $p(y|\vx)=\sN(y|100\cdot x_1, 1)$, that is, $\ry$ depends on $\rvx$ only through its first coordinate $\rx_1$. 

Firstly, we calculate $D_2(P_\lambda || Q)$ and $\text{ESS}^*(P_\lambda, Q)$ as functions of $d$ and then simulate how the predictive power of a decision tree regressor deteriorates as $d$ increases and $\text{ESS}^*(P_\lambda, Q)$ decreases. We train the trees by minimizing the empirical error weighted by the true weighting function $w$ in the training set, also imposing a minimum of 10 samples per leaf as a regularization strategy. We choose to work with decision trees since they are fast to train and robust against irrelevant features. Thus, it is reasonable to expect that a great part of performance deterioration is not due to noisy features but because of small ESSs.

The first step to calculate $\text{ESS}^*(P_\lambda, Q)$ and $D_2(P_\lambda || Q)$ is to calculate $\text{exp}[D_2(P_\lambda || Q)]$:
\begin{align}
   \text{exp}&[D_2(P_\lambda || Q)]=\E_{\rvx \sim P_\lambda}\left[\frac{p_\lambda(\rvx)}{q(\rvx)}\right]\\[.5em]
   &=\E_{\rvx \sim P_\lambda}\left\{\frac{\text{exp}[-\frac{1}{2}(\rvx-\lambda \mathbf{1})^\top (\rvx-\lambda \mathbf{1})]}{\text{exp}[-\frac{1}{2}\rvx^\top \rvx]}\right\}\\[.5em]
   &=\text{exp}\left(-\frac{d \lambda^2 }{2}\right)\cdot \E_{\rvx \sim P_\lambda}\left[\text{exp}\left(\lambda \sum_{j=1}^d \rx_j\right)\right]\\[.5em]
   &=\text{exp}(d \lambda^2)
\end{align}

The last equality is true since $\text{exp}(\lambda \sum_{j=1}^d \rx_j) \sim \text{LogNormal}(d \lambda^2,d \lambda^2)$. Then, $D_2(P_\lambda || Q)=d \lambda^2$ and \linebreak $\text{ESS}^*(P_\lambda, Q)=\text{exp}(-d \lambda^2)$. 

\begin{figure*}[ht]
  \centering
  \includegraphics[width=1\textwidth]{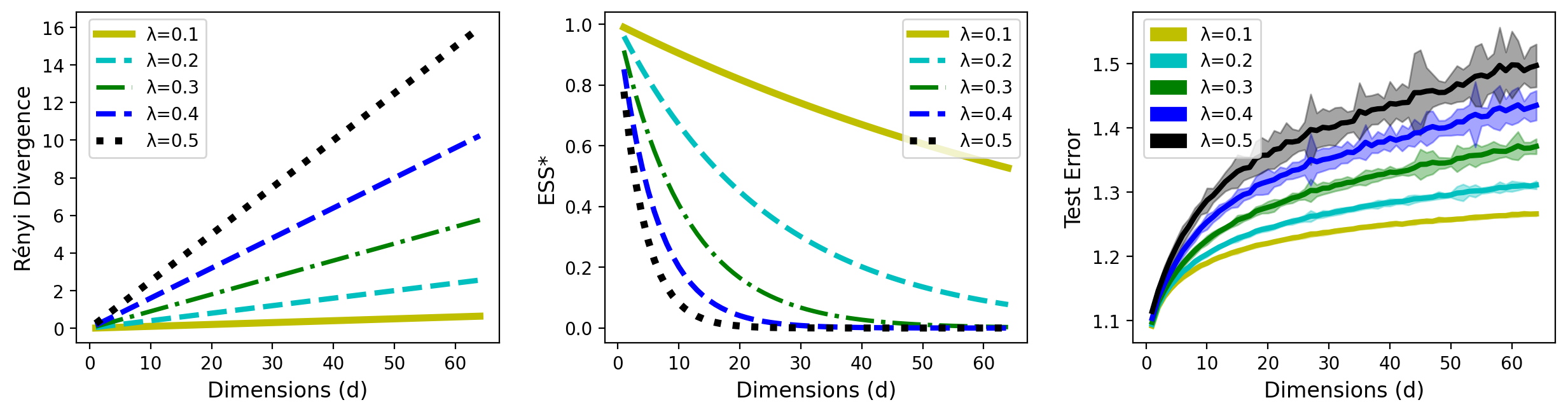}
  \caption{\footnotesize (i) We plot the Rényi Divergence of the target distribution $P_\lambda$ from the source distribution $Q$ as a function of the number of features. Both distributions are normal with the same covariance matrix but located $\sqrt{d\lambda^2}$ units apart from each other, i.e. the divergence also depends on $|\lambda|$; (ii) We plot the $\text{ESS}^*(P_\lambda, Q)$ as a function of $d$ and also varying $\lambda$. As expected, $\text{ESS}^*(P_\lambda, Q)$ exponentially decays in $d$ as long as the divergence is linearly related with $d$; (iii) In 50 simulations for each pair $(\lambda,d)$, we observe how decision trees' performances deteriorate as the divergence between domains grows and the ESS decreases.}
  \label{fig:toy}
\end{figure*}

Figure \ref{fig:toy} depicts the behavior of Rényi Divergence and $\text{ESS}^*(P_\lambda, Q)$ as functions of $d$. We also vary the value for $\lambda$. Given that $D_2(P_\lambda || Q)$ only depends on $|\lambda|$ and not on sign$(\lambda)$, we consider the case where $\lambda>0$. When $|\lambda|$ is bigger, the divergence between the source and target distributions also increases. Finally, to check how large $d$ affects performance of a regressor we, for each $d$, (i) sample 50 training and test sets of size $10^6$, (ii) train the trees on the training set minimizing the weighted empirical error and (iii) assess the regressors on the test sets. The third plot of Figure \ref{fig:toy} represents the average root-mean-square test error ($\pm$ standard deviation). Clearly the regressor deteriorates as the divergence between domains grows and the ESS decreases.

\section{The use of dimensionality reduction/feature selection to make effective sample size bigger}\label{sec:dim_red}

In this section, we present dimensionality reduction and feature selection as ways to obtain a bigger effective sample size. The two main results of this section are given by Theorems \ref{thm:div3} and \ref{thm:div4}. We show that linear dimensionality reduction and feature selection, under some conditions, decrease Rényi divergence between the target and source probability distributions, leading to a bigger effective sample size. This result accounts for the dimensionality reduction or feature selection's randomness; that is, the transformation can depend on data in some specific ways. 

To arrive at our main results, we first show the intermediate result given by Lemma \ref{lem:div2}. In the following result, $\mA$ represents a constant dimensionality reduction matrix and the vector $\vb$ represents a translation in data before dimensionality reduction, which is common when performing principal components analysis (PCA) \cite{hastie2009elements}, for example. When there is no need for considering a translation, we just can adopt $\vb=\mathbf{0}$. Also, $\mA$ can represent a feature selector, as we explain in the coming paragraphs.

\begin{lem}\label{lem:div2}
Consider (i) two absolutely continuous random vectors $\rvx \sim Q_\rvx$  and  $\rvx' \sim P_\rvx$ of size $d\geq2$, $D_2(P_{\rvx} || Q_{\rvx})<\infty$, (ii) a nonrandom constant vector $\vb \in \R^d$, and (iii) a nonrandom constant matrix $\mA \in \R^{d' \times d}$ with rank $d'$ (and $d' \leq d$). Suppose $Q_{\rvx}$ and $P_{\rvx}$ measure events in $\sX \subseteq \R^d$, $d\geq2$, and have probability density functions $q_{\rvx}$ and $p_{\rvx}$, such that $\text{support}(p_{\rvx}) \subseteq \text{support}(q_{\rvx})$. Also, assume $\mA (\rvx-\vb) \sim Q_{\mA (\rvx-\vb)}$ and $\mA (\rvx'-\vb) \sim P_{\mA (\rvx-\vb)}$. Then 
\begin{align}
D_2(P_{\rvx} || Q_{\rvx})  \geq  D_2(P_{\mA(\rvx-\vb)} || Q_{\mA(\rvx-\vb)})
\end{align}
And, consequently,
\begin{align}
 \textup{ESS}^*(P_{\mA(\rvx-\vb)},Q_{\mA(\rvx-\vb)}) \geq \textup{ESS}^*(P_{\rvx},Q_{\rvx})
\end{align}
\end{lem}

The proof can be found in the appendix. 

Although Lemma \ref{lem:div2} gives us a way out in cases which the dimensionality reduction is not random, this case is not realistic. We know that, in practice, $ \mA $ and $ \vb $ are obtained using data. 

In the next results, linear dimensionality reduction and feature selection are represented by the random matrix $ \rmA $. If we assume in advance that $ \rmA $ is absolutely continuous, then it represents an ordinary dimensionality reduction matrix. On the other hand, if $ \rmA $ is composed of zeros except for a single entry in each of its columns, which is given by one, then it represents a feature selector. Also, we can consider a random data translator $\rvb$ instead of the deterministic $\vb$.

\begin{thm}[Linear dimensionality reduction]\label{thm:div3}\linebreak 
Firstly, consider the training random samples of absolutely continuous vectors $\{\rvx_i\}_{i=1}^n \overset{iid}{\sim} Q_{\rvx}$ and an absolutely continuous random vector from target domain $\rvx' \sim P_{\rvx}$. Assume $Q_{\rvx}$ and $P_{\rvx}$ measure events in $\sX \subseteq \R^d$, $d\geq2$, and have probability density functions $q_{\rvx}$ and $p_{\rvx}$, such that $\text{support}(p_{\rvx}) \subseteq \text{support}(q_{\rvx})$. Also, assume that $D_2(P_{\rvx} || Q_{\rvx})<\infty$. Secondly, consider an absolutely continuous random vector $\rvb \in \R^{d}$ and an absolutely continuous random matrix $\rmA \in \R^{d' \times d}$, rank$(\mA)=d'$, jointly distributed according to the p.d.f. $p_{\rvb, \rmA}$, such that $(\rvb,\rmA), \rvx_i,$ and $\rvx'$ are pairwise independent, for every $i \in [n]$. Assume that $d' \leq d$. Suppose $\rmA(\rvx_i-\rvb) \sim Q_{\rmA(\rvx-\rvb)}$ and $\rmA(\rvx'-\rvb) \sim P_{\rmA(\rvx-\rvb)}$, for every $i \in [n]$, then 
\begin{align}
D_2(P_{\rvx} || Q_{\rvx})  \geq  D_2(P_{\rmA (\rvx-\rvb)} || Q_{\rmA(\rvx-\rvb)})
\end{align}
And, consequently,
\begin{align}
 \textup{ESS}^*(P_{\rmA(\rvx-\rvb)},Q_{\rmA(\rvx-\rvb)}) \geq \textup{ESS}^*(P_{\rvx},Q_{\rvx})
\end{align}
\end{thm}

The proof can be found in the appendix. 

Theorem \ref{thm:div3} tells us that a dimensionality reduction procedure before performing covariate shift adaptation increases the population effective sample size. It is important to state that Theorem \ref{thm:div3} also holds when disconsidering $\rvb$ and the proof's adaptation is straightforward. In that case, we would have that $D_2(P_{\rvx} || Q_{\rvx})  \geq  D_2(P_{\rmA\rvx} || Q_{\rmA\rvx})$ and $\textup{ESS}^*(P_{\rmA\rvx},Q_{\rmA\rvx}) \geq\textup{ESS}^*(P_{\rvx},Q_{\rvx})$.

Next, in Theorem \ref{thm:div4}, we state a result regarding feature selection.

\begin{thm}[Feature selection]\label{thm:div4}\linebreak 
Firstly, consider the training random samples of absolutely continuous vectors $\{\rvx_i\}_{i=1}^n \overset{iid}{\sim} Q_{\rvx}$ and an absolutely continuous random vector from target domain $\rvx' \sim P_{\rvx}$. Assume $Q_{\rvx}$ and $P_{\rvx}$ measure events in $\sX \subseteq \R^d$, $d\geq2$, and have probability density functions $q_{\rvx}$ and $p_{\rvx}$, such that $\text{support}(p_{\rvx}) \subseteq \text{support}(q_{\rvx})$. Also, assume that $D_2(P_{\rvx} || Q_{\rvx})<\infty$. Secondly, consider a discrete random matrix $\rmA \in \R^{d' \times d}$, that represents a feature selector with rank$(\mA)=d'$, distributed according to the probability mass function (p.m.f.) $p_{\rmA}$, such that $\rmA, \rvx_i,$ and $\rvx'$ are pairwise independent, for every $i \in [n]$. Assume that $d' \leq d$. Suppose $\rmA\rvx_i \sim Q_{\rmA\rvx}$ and $\rmA\rvx' \sim P_{\rmA\rvx}$, for every $i \in [n]$, then 
\begin{align}
D_2(P_{\rvx} || Q_{\rvx})  \geq  D_2(P_{\rmA\rvx} || Q_{\rmA\rvx})
\end{align}
And, consequently,
\begin{align}
 \textup{ESS}^*(P_{\rmA\rvx},Q_{\rmA\rvx}) \geq \textup{ESS}^*(P_{\rvx},Q_{\rvx})
\end{align}
\end{thm}

The proof can be found in the appendix. 

Theorems \ref{thm:div3} and  \ref{thm:div4} hold when the data used to obtain $\rmA$ and $\rvb$ do not depend on training data that will be used to train the supervised models or data points that represent the target domain we want to make generalizations for. That does not mean we cannot use some portion of the dataset to obtain $\rmA$ and $\rvb$, but it only means the results are not valid for those specific used data points, being from source or target domains. 

Before closing this section, it is worth mentioning three points. Firstly, at the same time dimensionality reduction/feature selection solve the problem of low effective sample sizes, it might impose other problems. For example, when performing principal components analysis (PCA) \cite{hastie2009elements} for dimensionality reduction, it is not guaranteed the method will not discard useful information for the supervised task. Also, it is not even possible to ensure the covariate shift main assumption, that the conditional distribution of the labels are the same in source and target domains, still holds. In this direction, \citet{stojanov2019low} offers a clever solution to overcome these specific problems, applying sufficient dimension reduction (SDR), which is a supervised method, to reduce dimensions. Secondly, given that $\rmA$ and $\rvb$ are random quantities\footnote{This is not true when $\rmA$ and $\rvb$ are fixed.}, $\{\rmA(\rvx_i-\rvb)\}_{i=1}^n$ or $\{\rmA\rvx_i\}_{i=1}^n$ may not form independent samples, even when $\rvx_i \independent (\rmA,\rvb), \forall i \in [n]$, and $\{\rvx_i\}_{i=1}^n \overset{iid}{\sim} Q_{\rvx}$. If samples are not independent, then the results presented in Section \ref{sec:ess} might not hold. Finally, it is true that the results presented in this section can be extended to include more general dimensionality reduction transformations, i.e. non-linear transformations, and the validity of other transformations might be proven using the Data Processing Inequality \cite{van2014renyi}. Unfortunately, exploring the two last points is beyond the scope of the present paper and might be treated in future work.

\section{Numerical experiments with real data}\label{sec:exp}


In this section, we present regression and classification experiments in which we perform feature selection before covariate shift adaptation. When designing the experiments, we choose to work with the least possible number of assumptions, searching for evidence that the theoretical results presented so far can be extended to more general cases, which will be treated in future work. Namely, we did not assume (i) the true importance function is always known, (ii) that training data is independent of the feature selector, and (iii) that training data are formed with independent data points after the feature selection procedure. 

For the following experiments, 10 regression datasets with no missing values were selected\footnote{From \url{www.dcc.fc.up.pt/~ltorgo/Regression/DataSets.html} and \url{https://archive.ics.uci.edu/ml/datasets.php}.}. Each experiment consisted of (i) introducing covariate shift\footnote{Similarly to previous research, e.g., \cite{huang2007correcting,reddi2015doubly,wang2017extreme,stojanov2019low}.}, (ii) estimating the weights, (iii) correcting the shift by the importance weighting method, and finally (iv) assessing the performance of the predictors and the effective sample size. We also studied the classification case by binarizing the target variables using their medians as a threshold. We use the same datasets for both regression and classification experiments to make comparisons easier. For each one of the $10$ datasets, we repeated the following preprocessing steps: (i) we kept up to 8,000 data points per dataset\footnote{The datasets “Abalone,” “Delta Ailerons,” and “Wine Quality” had 4177, 7129, and 6497 data points, respectively. All the others were undersampled to have 8,000 data points.}, (ii) generated new features using independent standard gaussian noise and (iii) standardized each column in every dataset. By augmenting the dataset to $32$ features using noise, we can explore a scenario in which performance deterioration is mainly due to small effective sample sizes. We give more details on this point in the next paragraph.

The following procedure is used to create divergent training and test sets after the preprocessing steps. For each of the datasets, we sampled a sequence of vectors uniformly from $[-1,1]^d$. We projected the data points onto the subspace generated by each vector, resulting in only one feature $\vx_i^{(j)}$ per sample $i$ for each subspace/simulation $j$. For each $\vx_i^{(j)}$, we calculated the score $s_{ij}=\Phi\big( [x_i^{(j)}-\text{median}(\vx^{(j)})]/\sigma_j\big)$, which is the probability that the data point $i$ from simulation $j$ is in the training set. According to that score, we randomly allocated each data point in either the training or test set in simulation $j$. The constant $\sigma_j$ was adjusted until the empirical effective sample size, as defined in Section \ref{sec:ess}, is less than $0.01$. Following this procedure, the training and test sets are approximately of the same sizes in each simulation $j$. Then, we fit two decision trees for each of the training/test sets: one in the training set and one in a subset of the test set. Then, we tested both decision trees in the unused portion of the test set and compared their performance according to the mean squared error for regression and classification error (1 - accuracy) for classification. We selected the 75 simulations\footnote{From the total of 7,200 simulations.} in which decision trees trained in the test sets did best, relatively to the training set trees. We chose decision trees because they are fast to train and robust against irrelevant features. Thus, the noisy features added in the datasets are not likely to directly affect predictive power but only by making the effective sample size smaller. It is important to state that, during the whole experimenting phase, decisions trees were 2-fold cross-validated in order to choose the minimum number of samples per leaf\footnote{More details on hyperparameter tuning can be found in the appendix}.

For the feature selection step, we were inspired by \citet{stojanov2019low} and the idea of Sufficient Dimension Reduction \cite{suzuki2010sufficient}, which is a supervised approach to dimensionality reduction and feature selection, contrasting to Principal Component Analysis, for example. Supervised approaches to dimensionality reduction and feature selection are preferable since we are able to keep important information for a supervised task performed afterwards. Using training data, we apply a combination of the methods described by \citet{lan2006estimating,eirola2014variable} and the \textit{Forward Selection} algorithm \cite{guyon2003introduction}. The approach uses gaussian mixture models (GMMs) to estimate, using the whole training set, the mutual information between a subset of features and the target variable. In this case, the number of GMMs' components are chosen evenly splitting the training data and performing a simple holdout set hyperparameter tuning phase\footnote{More details on hyperparameter tuning can be found in the appendix}. After training the GMMs, the procedure follows these steps: we start by choosing the feature that has the largest estimated mutual information with the target variable, and, at each subsequent step, we select the feature that marginally maximizes the estimated mutual information of target variable and selected features. We repeat the process until we reach a stop criteria. Our stopping criteria is that we should stop selecting features when the marginal improvement in the empirical mutual information is less than $1\%$ relative to the last level or when we select the first $15$ features. An implementation of the feature selection method is available in the Python package \textit{InfoSelect}\footnote{See \url{https://github.com/felipemaiapolo/infoselect} or \url{https://pypi.org/project/infoselect/}}. 

To estimate the weighting function for covariate shift adaptation, we use the probabilistic classification approach \cite{sugiyama2012density2,sugiyama2012machine} with a logistic regression model combined with a quadratic polynomial expansion of the original features. We choose to work with this approach since it is simple and fast to implement, besides being promising for high-dimensional settings. Others approaches are possible though \cite{sugiyama2012machine}. In order to prepare the data for training the logistic regression model, we first append the whole training set and randomly select rows ($80\%$) from the test set, and create the artificial labels for both groups. Then, we randomly/evenly split that dataset in order to choose the best value for the $l1$ regularization hyperparameter of the logistic regression, using the simple holdout validation approach\footnote{More details on hyperparameter tuning can be found in the appendix}. After getting the optimal values for the hyperparameter, we train a final model using the whole appended dataset.

In the experiments, we work with four training scenarios. In the first one, we use the whole set of features and no weighting method. In the second one, we use the entire set of features and importance weighting combined with the “true” weights $(1-s_{ij})/s_{ij}$. In the third, we use the whole set of features and estimated weights using the probabilistic classification approach. In the fourth scenario, we use only selected features and estimated weights using the probabilistic classification approach. Comparing the four scenarios enables us to see how importance weighting may fail in high-dimensional settings due to low ESS, even when we know the “true” weighting function.

Table \ref{tab:nvars} shows, for each one of the employed datasets, (i) the original number of features, (ii) the augmented number of features, (iii) the average number ($\pm$ standard deviation) of selected features for the regression and (iv) classification experiments.

\begin{table}[ht] 
 \centering 
 \caption{Average Numbers of features ($\pm$ standard deviation) - in this table we compare the numbers of original, augmented and selected features for regression (reg.) and classification (class.) tasks. It is possible to note that, on average, we select small subsets of features, even smaller than the original sets.} 
 \label{tab:nvars}%
 \resizebox{\columnwidth}{!}{
 \setlength\tabcolsep{2.5pt}
 \begin{tabular}{c|cc|cc} 
 \hline 
 Dataset & Original & Augmented & Selected (Reg.) & Selected (Class.) \\ 
 \hline 

abalone & $ 7 $ & $ 32 $ & $ 4.19  \pm  1.26 $ & $ 9.87  \pm  5.64 $  \\ 
ailerons & $ 40 $ & $ 40 $ & $ 5.16  \pm  0.54 $ & $ 3.79  \pm  0.64 $  \\ 
bank32nh & $ 32 $ & $ 32 $ & $ 10.00  \pm  1.82 $ & $ 13.91  \pm  0.61 $  \\ 
cal housing & $ 8 $ & $ 32 $ & $ 5.29  \pm  1.29 $ & $ 7.45  \pm  4.92 $  \\ 
cpu act & $ 21 $ & $ 32 $ & $ 9.88  \pm  1.20 $ & $ 2.56  \pm  0.72 $  \\ 
delta ailerons & $ 5 $ & $ 32 $ & $ 3.16  \pm  0.49 $ & $ 3.75  \pm  0.63 $  \\ 
elevators & $ 18 $ & $ 32 $ & $ 7.97  \pm  1.11 $ & $ 13.08  \pm  2.16 $  \\ 
fried delve & $ 10 $ & $ 32 $ & $ 4.45  \pm  0.50 $ & $ 5.00  \pm  0.00 $  \\ 
puma32H & $ 32 $ & $ 32 $ & $ 1.88  \pm  0.32 $ & $ 14.00  \pm  0.00 $  \\ 
winequality & $ 11 $ & $ 32 $ & $ 9.60  \pm  1.02 $ & $ 14.00  \pm  0.00 $  \\ 

 \hline 
 \end{tabular}%
 } 
 \end{table}%

\begin{figure}[ht]
 \centering
 \includegraphics[width=.4\textwidth]{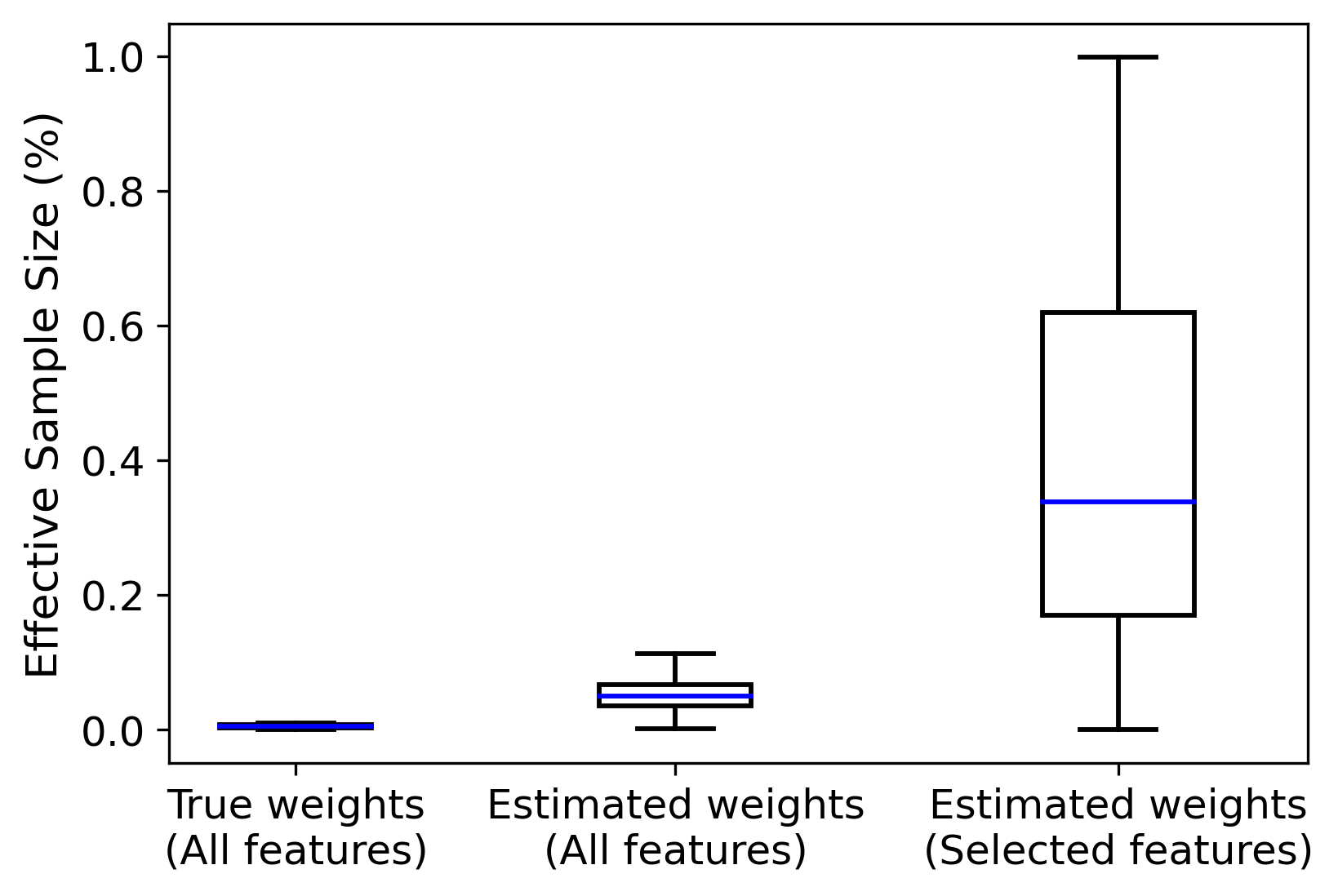}
 \caption{ Effective Sample Size distributions across all experiments. Notice higher ESSs can be achieved by a prior feature selection stage.}
 \label{fig:box}
\end{figure}

In Figure \ref{fig:box}, one can see the distribution of effective sample sizes in all the weighted approaches, calculated in the entire set of experiments. It is possible to notice how small the ESSs can be by adopting the pure weighting strategy. The feature selection step allows bigger ESSs.

In Table \ref{tab:performance}, we see the average test errors ($\pm$ standard deviation). To compute the errors, we use the test set portion ($20\%$) not used to train the importance function. The errors reported are the mean squared error and classification error relative to the first scenario. From Table \ref{tab:performance}, it is noticeable that our feature selection approach and posterior weighting systematically outperforms all the other benchmarks, especially the pure weighting method when the whole set of features is used. Even the benchmarks that used true weights are often beaten by large margins. That suggests that the degradation in the model performances is mainly due to small effective sample sizes instead of difficulties estimating the weighting function.

Through our experiments, we were able to verify that the feature selection stage tends to increase the effective sample size, consequently allowing better performance of supervised methods.

\begin{table*}[ht] 
 \centering 
 \caption{ Average Test Errors ($\pm$ std. deviation) - here we compared the predictive performance of decision trees in the test set of 75 different simulations for each dataset. We have four basic scenarios: (i) whole set of features and no weighting method; (ii) whole set of features and use of “true" weights; (iii) whole set of features and estimated weights; (iv) selected features and estimated weights. The numbers reported are the mean squared error and classification error averages and their std. deviations. All the results were normalized w.r.t. the first scenario.} 
 \label{tab:performance}%
 \resizebox{350pt}{!}{ 
 \setlength\tabcolsep{3.5pt}
 \begin{tabular}{cc|ccc|c} 
 \hline 
 \multicolumn{1}{c}{} & \multicolumn{1}{c}{} & \multicolumn{3}{c}{All features} & \multicolumn{1}{c}{Selected features}  \\ 
 \hline 
  & Dataset & Unweighted & True weights & Estimated weights  & Estimated weights  \\ 

\hline 
 \multirow{10}{*}{\rotatebox[origin=c]{90}{Regression}} & abalone
& $ 1.00 $ & $ 1.42  \pm  0.24 $ & $ 1.25  \pm 0.19 $ & $ \mathbf{0.92  \pm  0.07} $ \\ 
& ailerons
& $ 1.00 $ & $ 1.01  \pm  0.13 $ & $ 0.99  \pm 0.11 $ & $ \mathbf{0.87  \pm  0.11} $ \\ 
& bank32nh
& $ 1.00 $ & $ 1.29  \pm  0.14 $ & $ 1.20  \pm 0.11 $ & $ \mathbf{0.98  \pm  0.06} $ \\ 
& cal housing
& $ 1.00 $ & $ 1.50  \pm  0.24 $ & $ 1.35  \pm 0.20 $ & $ \mathbf{0.84  \pm  0.09} $ \\ 
& cpu act
& $ 1.00 $ & $ 0.52  \pm  0.55 $ & $ 0.55  \pm 0.59 $ & $ \mathbf{0.15  \pm  0.21} $ \\ 
& delta ailerons
& $ 1.00 $ & $ 1.39  \pm  0.18 $ & $ 1.25  \pm 0.12 $ & $ \mathbf{0.92  \pm  0.06} $ \\ 
& elevators
& $ 1.00 $ & $ 1.10  \pm  0.15 $ & $ 1.05  \pm 0.13 $ & $ \mathbf{0.85  \pm  0.15} $ \\ 
& fried delve
& $ 1.00 $ & $ 1.60  \pm  0.22 $ & $ 1.40  \pm 0.15 $ & $ \mathbf{0.90  \pm  0.11} $ \\ 
& puma32H
& $ \mathbf{1.00} $ & $ 2.24  \pm  1.18 $ & $ 1.45  \pm 0.22 $ & $ 1.77  \pm  2.42 $ \\ 
& winequality
& $ 1.00 $ & $ 1.31  \pm  0.12 $ & $ 1.24  \pm 0.11 $ & $ \mathbf{0.97  \pm  0.04} $ \\ 
\hline 
 \multirow{10}{*}{\rotatebox[origin=c]{90}{Classification}} & abalone
& $ \mathbf{1.00} $ & $ 1.29  \pm  0.19 $ & $ 1.22  \pm 0.16 $ & $ 1.05  \pm  0.15 $ \\ 
& ailerons
& $ 1.00 $ & $ 1.03  \pm  0.27 $ & $ 1.01  \pm 0.20 $ & $ \mathbf{0.86  \pm  0.13} $ \\ 
& bank32nh
& $ \mathbf{1.00} $ & $ 1.25  \pm  0.13 $ & $ 1.20  \pm 0.13 $ & $ \mathbf{1.00  \pm  0.09} $ \\ 
& cal housing
& $ 1.00 $ & $ 1.43  \pm  0.23 $ & $ 1.36  \pm 0.19 $ & $ \mathbf{0.87  \pm  0.14} $ \\ 
& cpu act
& $ 1.00 $ & $ 1.09  \pm  0.16 $ & $ 1.06  \pm 0.16 $ & $ \mathbf{0.99  \pm  0.15} $ \\ 
& delta ailerons
& $ 1.00 $ & $ 1.38  \pm  0.40 $ & $ 1.25  \pm 0.31 $ & $ \mathbf{0.84  \pm  0.12} $ \\ 
& elevators
& $ 1.00 $ & $ 1.07  \pm  0.15 $ & $ 1.04  \pm 0.14 $ & $ \mathbf{0.89  \pm  0.13} $ \\ 
& fried delve
& $ 1.00 $ & $ 1.34  \pm  0.22 $ & $ 1.22  \pm 0.18 $ & $ \mathbf{0.85  \pm  0.09} $ \\ 
& puma32H
& $ \mathbf{1.00} $ & $ 1.73  \pm  0.59 $ & $ 1.22  \pm 0.18 $ & $ 1.10  \pm  0.42 $ \\ 
& winequality
& $ \mathbf{1.00} $ & $ 1.20  \pm  0.13 $ & $ 1.13  \pm 0.10 $ & $ 1.07  \pm  0.10 $ \\ 

 \hline 
 \end{tabular}%
 } 
 \end{table*}%

\section{Conclusion}


In this paper, we have made two main contributions. The first is that we explicitly and formally connected three key concepts in the context of covariate shift adaptation: (i) effective sample size, (ii) dimensionality of data, and (iii) generalization of a supervised model. Since, to the best of our knowledge, there is no unified and rigorous view on how the three key concepts connect to each other, we consider this to be the first contribution of the paper. The second contribution of the paper is that we show dimensionality reduction or feature selection, even considering data dependent mappings, corrects small effective sample sizes by making the source and target distributions less divergent. This suggests that it is a good practice to perform dimensionality reduction or feature selection before covariate adaptation. We also present numerical experiments using real and artificial data to complement our theoretical results.

Regarding possible future research paths and improvements, we point to Sections \ref{sec:ess} and \ref{sec:dim_red}. Concerning Section \ref{sec:ess}, perhaps the three most relevant points to be considered for future research relate to the following assumptions: the first one is assuming the importance function is known up to a constant, the second is assuming the sample ESS is close to its population version, and the third is assuming independent samples. While the first hardly applies in practice, the second may hold in many situations, and the third could be relaxed to include dependent samples, thus solving one of the problems discussed in Section \ref{sec:dim_red}. Considering Section \ref{sec:dim_red}, we think there is one main point to be explored in future work, which is extending the theorems to include more general transformations, i.e., non-linear or training data dependent transformations. Said that, future work and improvements of this paper could focus on relaxing assumptions or exploring cases in which they are valid.

\section{Computing Infrastructure}

All the experiments were carried out using a Google Cloud Platform's (GCP) Virtual Machine with 96 vCPUs and 86.4 GB of memory. All the experiments took around 4h to run.

\section{Declarations}

\subsection{Conflicts of interest/Competing interests}
The authors have no conflicts of interest to declare that are relevant to the content of this article.

\subsection{Funding}
We gratefully acknowledge financial support from Conselho Nacional de Desenvolvimento Científico
and Tecnológico (CNPq) and from the Advanced Institute for Artificial Intelligence (AI2), Brazil. Felipe Maia Polo was supported by the two institutions during his master's degree while writing this work.

\subsection{Availability of data and material}
All the datasets used are open datasets and are downloaded while running the code.

\subsection{Code availability}
The code and material used can be found online\footnote{\url{https://github.com/felipemaiapolo/ess_dimensionality_covariate_shift}}. Also, we have made our feature selection implementation available as a Python package called \textit{InfoSelect}\footnote{\url{https://github.com/felipemaiapolo/infoselect}}.


%
%

\bibliographystyle{unsrtnat}
\bibliography{bibliography.bib}

\newpage

\onecolumn
\section{Appendix}

\subsection{Proofs and Derivations}

\subsubsection{Proof of Theorem \ref{thm:ess}}
\begin{proof}
Assume the hypothesis stated are valid. Being $c\neq 0$ a real constant, see we can re-wright the ESS as follows:
\begin{align*}
 \widehat{\textup{ESS}}_n(P_\rvx, Q_\rvx)&=\frac{(\sum_{i=1}^n \rw_i)^2}{n\sum_{i=1}^n \rw_i^2}=\frac{\left[\sum_{i=1}^n c\cdot \frac{p_{\rvx}(\rvx_i)}{q_{\rvx}(\rvx_i)}\right]^2}{n\sum_{i=1}^n  \left[c\cdot \frac{p_{\rvx}(\rvx_i)}{q_{\rvx}(\rvx_i)}\right]^2}=\frac{\left[\frac{1}{n}\sum_{i=1}^n  \frac{p_{\rvx}(\rvx_i)}{q_{\rvx}(\rvx_i)}\right]^2}{\frac{1}{n}\sum_{i=1}^n  \left[\frac{p_{\rvx}(\rvx_i)}{q_{\rvx}(\rvx_i)}\right]^2}\\[.5em]
\end{align*}

Then, by the Strong Law of Large Numbers and almost-sure convergence properties \cite{roussas1997course}, we verify that $\widehat{\textup{ESS}}_n(P_\rvx, Q_\rvx) \xrightarrow[]{a.s.} \frac{\E_{\rvx \sim Q_\rvx}\left[\frac{p_\rvx(\rvx)}{q_\rvx(\rvx)} \right]^2}{\E_{\rvx \sim Q_\rvx}\left[\left(\frac{p_\rvx(\rvx)}{q_\rvx(\rvx)}\right)^2 \right]} \text{ when } n \rightarrow \infty$. To complete the proof, we state the following
\begin{align*}
 \frac{\E_{\rvx \sim Q_\rvx}\left[\frac{p_\rvx(\rvx)}{q_\rvx(\rvx)} \right]^2}{\E_{\rvx \sim Q_\rvx}\left[\left(\frac{p_\rvx(\rvx)}{q_\rvx(\rvx)}\right)^2 \right]} &=\frac{1}{\E_{\rvx \sim P_\rvx}\left[\frac{p_\rvx(\rvx)}{q_\rvx(\rvx)} \right]}=\frac{1}{\textup{exp}\left[D_2(P_\rvx || Q_\rvx)\right]}=\text{ESS}^*(P_\rvx, Q_\rvx)
\end{align*}
\end{proof}

\subsubsection{Proof of Theorem \ref{thm:div1}}
\begin{proof}
Assume the hypothesis are valid and let $d_2(P_{\rvx_1,\rvx_2} || Q_{\rvx_1,\rvx_2})$ = $\text{exp}[D_2(P_{\rvx_1,\rvx_2} || Q_{\rvx_1,\rvx_2})]$. See that:
\begin{align*}
 d_2(P_{\rvx_1,\rvx_2} || Q_{\rvx_1,\rvx_2})&= \E_{P_{\rvx_1,\rvx_2}} \left[\frac{p_{\rvx_1,\rvx_2}(\rvx_1,\rvx_2)}{q_{\rvx_1,\rvx_2}(\rvx_1,\rvx_2)}\right] = \E_{P_{\rvx_1}} \left[\frac{p_{\rvx_1}(\rvx_1)}{q_{\rvx_1}(\rvx_1)}\cdot \E_{P_{\rvx_2|\rvx_1}} \left[\frac{p_{\rvx_2|\rvx_1}(\rvx_2|\rvx_1)}{q_{\rvx_2|\rvx_1}(\rvx_2|\rvx_1)}\right]\right] =\\[.5em]
 &=\E_{P_{\rvx_1}} \left[\frac{p_{\rvx_1}(\rvx_1)}{q_{\rvx_1}(\rvx_1)}\cdot d_2(P_{\rvx_2|\rvx_1} || Q_{\rvx_2|\rvx_1})\right] \geq \E_{P_{\rvx_1}}   \left[\frac{p_{\rvx_1}(\rvx_1)}{q_{\rvx_1}(\rvx_1)}\right]=d_2(P_{\rvx_1} || Q_{\rvx_1})\\[.5em]
\end{align*}

Where the inequality is obtained by the fact that the exponential of the Rényi Divergence must be greater or equals one. To complete the proof, see that $\textup{ESS}^*(P_{\rvx_1,\rvx_2}, Q_{\rvx_1,\rvx_2})=d_2(P_{\rvx_1,\rvx_2} || Q_{\rvx_1,\rvx_2})^{-1}$. Therefore
\begin{align*}
    \textup{ESS}^*(P_{\rvx_1}, Q_{\rvx_1}) \geq \textup{ESS}^*(P_{\rvx_1,\rvx_2}, Q_{\rvx_1,\rvx_2})
\end{align*}
\end{proof}

\subsubsection{Proof of Lemma \ref{lem:div2}}
\begin{proof}
If $d=d'$, the result is direct, considering the arguments used by \cite{qiao2010study} to prove\footnote{Even though $D_2$ is not an f-divergence, the thoughts presented by \cite{qiao2010study} in their proof can readily be applied in this case. Furthermore, we can write $ D_2(P_{\rvx} || Q_{\rvx})=\log(\chi^2(P_{\rvx} || Q_{\rvx})+1)$, where $\chi^2$ is a f-divergence \cite{sason2016f}. This is an another reason on why this is valid.\label{note1}} their Theorem 1, because $\mA$ represents an invertible linear (and differentiable) transformation. Otherwise, consider a full rank matrix $\mC=\begin{bmatrix}
\mA\\ 
\mB
\end{bmatrix} \in \R^{d \times d}$, where $\mB \in \R^{d-d',d}$. Given that $\mC$ is full rank, it represents an invertible linear (and differentiable) transformation. If $\mC (\rvx-\vb)=\begin{bmatrix}
\mA(\rvx-\vb)\\ 
\mB(\rvx-\vb)
\end{bmatrix} \sim Q_{\mC (\rvx-\vb)}$ and $\mC (\rvx'-\vb)=\begin{bmatrix}
\mA(\rvx'-\vb)\\ 
\mB(\rvx'-\vb)
\end{bmatrix}\sim P_{\mC (\rvx'-\vb)}$, then by the arguments used by \cite{qiao2010study} to prove\footref{note1} their Theorem 1, we have that $ D_2(P_{\rvx} || Q_{\rvx}) = D_2(P_{\mC (\rvx-\vb)} || Q_{\mC (\rvx-\vb)})$. Discarding $\mB(\rvx-\vb)$ and $\mB(\rvx'-\vb)$ from random vectors $\mC(\rvx-\vb)$ and $\mC(\rvx'-\vb)$, by Theorem \ref{thm:div1}, we have that
\begin{align*}
D_2(P_{\rvx} || Q_{\rvx}) \geq D_2(P_{\mA (\rvx-\vb)} || Q_{\mA (\rvx-\vb)})
\end{align*}

Therefore
\begin{align*}
  &\textup{ESS}^*(P_{\mA (\rvx-\vb)}, Q_{\mA (\rvx-\vb)})=\exp\left[-D_2(P_{\mA (\rvx-\vb)} || Q_{\mA (\rvx-\vb)}) \right]\geq \exp\left[-D_2(P_{\rvx} || Q_{\rvx}) \right] =\textup{ESS}^*(P_{\rvx}, Q_{\rvx})
\end{align*}
\end{proof}

\subsubsection{Proof of Theorem \ref{thm:div3}}
\begin{proof}
Firstly, we define $\rvv:=\rmA (\rvx_i-\rvb) \sim Q_{\rvv} \equiv Q_{\rmA (\rvx-\rvb)}$ and $\rvu:=\rmA (\rvx'-\rvb) \sim P_{\rvu} \equiv P_{\rmA (\rvx-\rvb)}$, for an arbitrary $i \in [n]$. Let $q_{\rvv}$ and $p_{\rvu}$ be probability density functions associated with distributions $Q_{\rvv}$ and $P_{\rvu}$. From Lemma \ref{lem:div2}, we know that  $D_2\left(P_{\rvu|\rvb=\vb,\rmA=\mA}||Q_{\rvv|\rvb=\vb,\rmA=\mA} \right) \leq D_2\left( P_{ \rvx} || Q_{ \rvx} \right), \forall\vb\in\R^d$, $\forall \mA \in \R^{d' \times d}$ such that rank$(\mA)=d'$. That statement implies the following:
\begin{align*}
  &D_2\left(P_{\rvu|\rvb=\vb,\rmA=\mA}||Q_{\rvv|\rvb=\vb,\rmA=\mA} \right) \leq D_2\left( P_{\rvx} || Q_{\rvx} \right)\Rightarrow \exp D_2\left(P_{\rvu|\rvb=\vb,\rmA=\mA}||Q_{\rvv|\rvb=\vb,\rmA=\mA} \right) \leq \exp D_2\left( P_\rvx || Q_\rvx \right)\Rightarrow\\[1em]
  &\Rightarrow \E_{p_{\rvb, \rmA}}\left[\exp D_2\left(P_{\rvu|\rvb,\rmA}||Q_{\rvv|\rvb,\rmA} \right)\right] \leq \exp D_2\left( P_\rvx || Q_\rvx \right)\Rightarrow\\[1em]
  &\Rightarrow \int p_{\rvb, \rmA}(\vb,\mA)\int p_{\rvu|\rvb,\rmA}(\vu|\vb,\mA)\frac{p_{\rvu|\rvb,\rmA}(\vu|\vb,\mA)}{q_{\rvv|\rvb,\rmA}(\vu|\vb,\mA)}\text{d}\vu\text{d}\vb\text{d}\mA \leq \exp D_2\left( P_\rvx || Q_\rvx \right)\Rightarrow\\[1em]
  &\Rightarrow \int p_{\rvu|\rvb,\rmA}(\vu|\vb,\mA)p_{\rvb, \rmA}(\vb,\mA)\frac{p_{\rvu|\rvb,\rmA}(\vu|\vb,\mA)}{q_{\rvv|\rvb,\rmA}(\vu|\vb,\mA)}\frac{p_{\rvb, \rmA}(\vb,\mA)}{p_{\rvb, \rmA}(\vb,\mA)}\text{d}\vu\text{d}\vb\text{d}\mA \leq \exp D_2\left( P_\rvx || Q_\rvx \right)\Rightarrow\\[1em]
  &\Rightarrow D_2\left(P_{\rvu,\rvb,\rmA}||Q_{\rvv,\rvb,\rmA} \right) \leq D_2\left( P_\rvx || Q_\rvx \right)\Rightarrow D_2\left(P_{\rmA (\rvx-\rvb)}||Q_{\rmA (\rvx-\rvb)} \right) = D_2\left(P_{\rvu}||Q_{\rvv} \right) \leq D_2\left(P_{\rvu,\rvb,\rmA}||Q_{\rvv,\rvb,\rmA} \right)\leq D_2\left( P_\rvx || Q_\rvx \right)\\[1em]
\end{align*}

The last step is due to Theorem \ref{thm:div1} (extending to random matrices). To complete the proof, we state the following:
\begin{align*}
  &\textup{ESS}^*(P_{\rmA (\rvx-\rvb)}, Q_{\rmA (\rvx-\rvb)})=\exp\left[-D_2(P_{\rmA (\rvx-\rvb)} || Q_{\rmA (\rvx-\rvb)}) \right]\geq \exp\left[-D_2(P_{\rvx} || Q_{\rvx}) \right] =\textup{ESS}^*(P_{\rvx}, Q_{\rvx})
\end{align*}
\end{proof}

\subsubsection{Proof of Theorem \ref{thm:div4}}
\begin{proof}
Firstly, we define $\rvv:=\rmA \rvx_i \sim Q_{\rvv} \equiv Q_{\rmA \rvx}$ and $\rvu:=\rmA \rvx' \sim P_{\rvu} \equiv P_{\rmA \rvx}$, for an arbitrary $i \in [n]$. Let $q_{\rvv}$ and $p_{\rvu}$ be probability density functions associated with distributions $Q_{\rvv}$ and $P_{\rvu}$. From Lemma \ref{lem:div2}, we know that  $D_2\left(P_{\rvu|\rmA=\mA}||Q_{\rvv|\rmA=\mA} \right) \leq D_2\left( P_{ \rvx} || Q_{ \rvx} \right)$, $\forall \mA \in \R^{d' \times d}$ such that rank$(\mA)=d'$. That statement implies the following:
\begin{align*}
  &D_2\left(P_{\rvu|\rmA=\mA}||Q_{\rvv|\rmA=\mA} \right) \leq D_2\left( P_{\rvx} || Q_{\rvx} \right)\Rightarrow \exp D_2\left(P_{\rvu|\rmA=\mA}||Q_{\rvv|\rmA=\mA} \right) \leq \exp D_2\left( P_\rvx || Q_\rvx \right)\Rightarrow\\[1em]
  &\Rightarrow \E_{p_\rmA}\left[\exp D_2\left(P_{\rvu|\rmA}||Q_{\rvv|\rmA} \right)\right] \leq \exp D_2\left( P_\rvx || Q_\rvx \right)\Rightarrow \sum_{\mA} p_\rmA(\mA)\int p_{\rvu|\rmA}(\vu|\mA)\frac{p_{\rvu|\rmA}(\vu|\mA)}{q_{\rvv|\rmA}(\vu|\mA)}\text{d}\vu \leq \exp D_2\left( P_\rvx || Q_\rvx \right)\Rightarrow\\[1em]
  &\Rightarrow \sum_{\mA} \int  p_{\rvu|\rmA}(\vu|\mA)p_\rmA(\mA)\frac{p_{\rvu|\rmA}(\vu|\mA)}{q_{\rvv|\rmA}(\vu|\mA)}\frac{p_\rmA(\mA)}{p_\rmA(\mA)}\text{d}\vu \leq \exp D_2\left( P_\rvx || Q_\rvx \right)\Rightarrow D_2\left(P_{\rvu,\rmA}||Q_{\rvv,\rmA} \right) \leq D_2\left( P_\rvx || Q_\rvx \right)\Rightarrow\\[1em]
  &\Rightarrow D_2\left(P_{\rmA \rvx}||Q_{\rmA \rvx} \right) = D_2\left(P_{\rvu}||Q_{\rvv} \right) \leq D_2\left(P_{\rvu,\rmA}||Q_{\rvv,\rmA} \right)\leq D_2\left( P_\rvx || Q_\rvx \right)\\[1em]
\end{align*}

Given the matrix $\mA$ represents a feature selector, it can only assume a finite number of values. Thus, the sum is given over a finite number of values of $\mA$. The last step is due to the Theorem \ref{thm:div1} (extending to random matrices). To complete the proof, we state the following:
\begin{align*}
  &\textup{ESS}^*(P_{\rmA \rvx}, Q_{\rmA \rvx})=\exp\left[-D_2(P_{\rmA \rvx} || Q_{\rmA \rvx}) \right]\geq \exp\left[-D_2(P_{\rvx} || Q_{\rvx}) \right] =\textup{ESS}^*(P_{\rvx}, Q_{\rvx})
\end{align*}
\end{proof}

\subsection{Experiments}

In the experiments section, we tune three hyperparameters: (i) $l1$ regularization parameter used to train the logistic regression model when estimating $w$, (ii) the minimum number of samples per leaf in each regression/classification tree, and (iii) number of GMM components. We use the Scikit-Learn \cite{pedregosa2011scikit} implementations to train the logistic regressions, regression/classification trees and GMMs. Firstly, we choose the $l1$ logistic regression regularization parameter $C$ from values in $[10^{-4},5]$, in order to minimize the log loss in a holdout dataset. Secondly, we choose the minimum number of samples per leaf in each regression/classification tree from values in $[5,15,25,40,50]$, in order to minimize the mean squared error or classification error within a 2-fold cross-validation procedure. Finally, we maximize the log-likelihood in a holdout dataset to choose the number of GMM components, varying the possible number of components within the list $[1,3,5,7,9,11,13,15]$.

\end{document}